\pdfoutput=1

\documentclass[11pt]{article}

\usepackage[]{acl}

\usepackage{times}
\usepackage{latexsym}
\usepackage[pdftex]{graphicx}
\usepackage{booktabs}
\usepackage{amsmath}

\usepackage[T1]{fontenc}

\usepackage[utf8]{inputenc}

\usepackage{microtype}
\usepackage{ulem}

\title{Dynamic Position Encoding for Transformers}

\author{Joyce Zheng\thanks{\hspace{2mm}Work done while Joyce Zheng was an intern at Huawei} \\
  Huawei Noah's Ark Lab\\
  \texttt{jy6zheng@uwaterloo.ca} \\
  \And
  Mehdi Rezagholizadeh \\
  Huawei Noah's Ark Lab \\
  \texttt{mehdi.rezagholizadeh@huawei.com}\\
  \AND
  Peyman Passban \thanks{\hspace{2mm}Work done while Peyman Passban was at Huawei.} \\
  Amazon \\
  \texttt{passban.peyman@gmail.com} \\}

\begin{document}
\maketitle
\begin{abstract}
Recurrent models have been dominating the field of neural machine translation (NMT) for the past few years. Transformers \citep{vaswani2017attention} have radically changed it by proposing a novel architecture that relies on a feed-forward backbone and self-attention mechanism. Although Transformers are powerful, they could fail to properly encode sequential/positional information due to their non-recurrent nature. To solve this problem, position embeddings are defined exclusively for each time step to enrich word information. However, such embeddings are fixed after training regardless of the task and word ordering system of the source and target languages. 
 
In this paper, we address this shortcoming by proposing a novel architecture with new position embeddings that take the order of the target words into consideration. Instead of using predefined position embeddings, our solution \textit{generates} new embeddings to refine each word's position information. Since we do not dictate the position of the source tokens and we learn them in an end-to-end fashion, we refer to our method as \textit{dynamic} position encoding (DPE). We evaluated the impact of our model on multiple datasets to translate from English to German, French, and Italian and observed meaningful improvements in comparison to the original Transformer.
\end{abstract}

\section{Introduction}
In statistical machine translation (SMT), the general task of translation consists of reducing the input sentence into smaller units (also known as statistical phrases), selecting an optimal translation for each unit, and placing them in the correct order \citep{koehn2009statistical}. The last step, which is also referred to as the reordering problem, is a great source of complexity and importance, which is handled with a variety of statistical as well as string- and tree-based solutions \citep{bisazza-federico-2016-surveys}. 

As evident in SMT, the structure and position of words within a sentence is crucial for accurate translation. The importance of such information can also be explored in NMT and for Transformers. Previous literature, such as \citet{chen2020explicit}, demonstrated that source input sentences enriched with target order information have the capacity to improve translation quality in neural models. They showed that position encoding seems to play a key role in translation and this motivated us to further explore this area. 

Since Transformers have a non-recurrent architecture, they could face problems when encoding sequential data. As a result, they require an explicit summation of the input embeddings with position encoding to provide information about the order of each word. However, this approach falsely assumes that the correct position of each word is always its original position in the source sentence. This interpretation might be true when only considering the source side, whereas we know from SMT \citep{bisazza-federico-2016-surveys,cui-etal-2016-lstm} that  arranging input words with respect to the order of their target pairs can lead to better results. 

In this work, we explore injecting target position information alongside the source words to enhance Transformers' translation quality. We first examine the accuracy and efficiency of a two-pass Transformer (2PT), which consists of a pipeline connecting two Transformers. The first Transformer reorders the source sentence and the second one translates the reordered sentences. Although this approach incorporates order information from the target language, it lacks end-to-end training and requires more resources than a typical Transformer. Accordingly, we introduce a better alternative, which effectively learns the reordered positions in an end-to-end fashion and uses this information with the original source sequence to boost translation accuracy. We refer to this alternative as \textit{Dynamic Position Encoding} (DPE).

Our contribution in this work is threefold:
\begin{itemize}
  \item First, we demonstrate that providing source-side representations with target position information improves translation quality in Transformers.
  
  \item We also propose a novel architecture, DPE, that efficiently learns reordered positions in an end-to-end fashion and incorporates this information into the encoding process.
  
  \item Finally, we use a preliminary two-pass design to show the importance of end-to-end learning in this problem.
  
\end{itemize}

\section{Background}\label{background}
\subsection{Pre-Reordering in Machine Translation}
\begin{figure}[h!]
\centering
\includegraphics[scale=0.37]{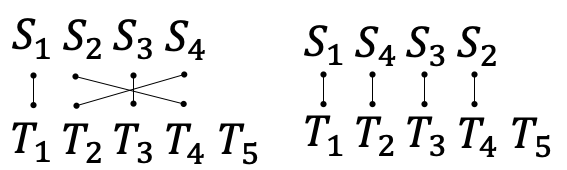}
\caption{\label{fig:1}The order of source words before (left-hand side) and after (right-hand side) pre-reordering. S$_i$ and T$_j$ show the $i$-th source and $j$-th target words, accordingly.}
\end{figure}

In standard SMT, pre-reordering is a well-known technique. Usually, the source sentence is reordered using heuristics such that it follows the word order of the target language. Figure \ref{fig:1} illustrates this concept with an imaginary example.

As the figure shows, the original alignment between the source ($S_i$) and target ($T_i$) words are used to define a new order for the source sentence. With the new order, the translation engine does not need to learn the relation between source and target ordering systems, as it directly translates from one position on the source side to the \textit{same position} on the target side. Clearly, this can significantly reduce the complexity of the translation task. The work of \citet{wang-etal-2007-chinese}, provides a good example of systems with pre-reordering, in which the authors studied this technique for the English--Chinese pair.

The concept of re-ordering does not necessarily need to be tackled prior to translation; in \citet{koehn-etal-2007-moses}, generated sequences are reviewed by a classifier after translation to correct the position of words that are placed in the wrong order. The entire reordering process can also be embedded into the decoding process \citep{Feng2013AdvancementsIR}. 

\subsection{Tackling the Order Problem in NMT}
Pre-reordering and position encoding are also common in NMT and have been investigated by various researchers. \citet{du2017pre} explored if recurrent neural models can benefit from pre-reordering. Their findings showed that these models might not require any order adjustments because the network itself was powerful enough to learn such mappings. \citet{kawara2020preordering}, unlike the previous work, studied the same problem and reported promising observations on the usefulness of pre-reordering. They used a transduction-grammar-based method with recursive neural networks and showed how impactful pre-reordering could be.

\citet{liu2020learning} followed a different approach and proposed modeling position encoding as a continuous dynamical system through a neural ODE. \citet{ke2020rethinking} investigated improving positional encoding by untying the relationship between words and positions. They suggested that there is no strong correlation between words and absolute positions, so they removed this noisy correlation. This form of separation has its own advantages, but by removing this relationship between words and positions from the translation process, they might lose valuable semantic information about the source and target sides.

\citet{DBLP:journals/corr/abs-1803-02155} explored a relative position encoding method by having the self-attention mechanism consider the distance between source words. \citet{garg2019jointly} incorporated target position information via multitasking where a translation loss was combined with an alignment loss that supervised one decoder head to learn position information. \citet{chen2020explicit} changed the Transformer architecture to incorporate order information at each layer. They selected a reordered target word position from the output of each layer and injected it into the next layer. This is the closest work to ours, so we consider it as our main baseline. 

\section{Methodology}\label{method}
\subsection{Two-Pass Translation for Pre-Reordering}\label{twopass}

Our goal is to boost translation quality in Transformers by injecting target order information into the source-side  encoding process. We propose dynamic position encoding (DPE) to achieve this, but before that, we discuss a preliminary two-pass Transformer (2PT) architecture to demonstrate the impact of order information in Transformers. 

\begin{figure} [h]
\centering
  \includegraphics[scale=0.5]{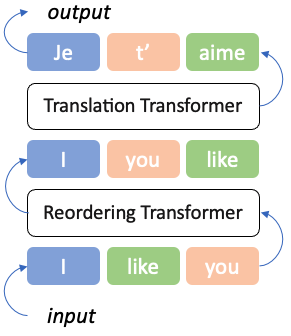}
\caption{\label{fig:2pe} The two-pass Transformer architecture. The input sequence is first re-ordered to a new and less complex form for the translation Transformer. Then, the translation Transformer uses the re-ordered input sequence to decode a target sequence.}
\end{figure}

The main difference between 2PT and DPE is that DPE is an end-to-end, differential solution whereas 2PT is a pipeline that connects two \textit{different} Transformers. They both work towards leveraging order information to improve translation but in different fashions. The 2PT architecture is illustrated in Figure \ref{fig:2pe}. 

2PT has two different Transformers. The first one is used for reordering purposes instead of translation. It takes source sentences and generates a reordered version of them, e.g. referring back to Figure \ref{fig:1}, if the input to the first Transformer is [$S_1$, $S_2$, $S_3$, $S_4$] the expected output from the first transformer is  [$S_1$, $S_4$, $S_3$, $S_2$]. We created a new corpus using {\fontfamily{pcr}\selectfont FastAlign} to train this reordering model \citep{dyer-etal-2013-simple}.\footnote{\url{https://github.com/clab/fast_align}} 

\begin{figure*} [ht]
\centering
\includegraphics[scale=0.5]{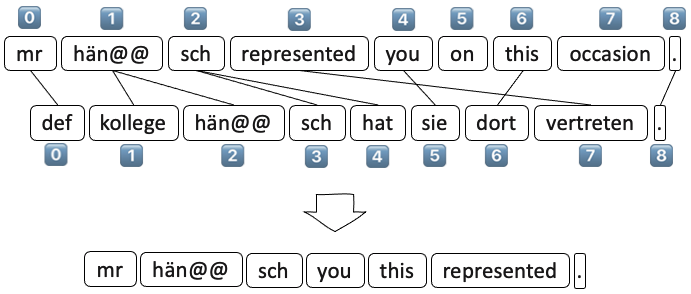}
\caption{\label{fig:fastal}{\fontfamily{pcr}\selectfont FastAlign}-based reordering using a sample sentence from our English--German dataset. Using word alignments, we generate a new reordered form from each source sentence as the new target sequence. We then use the pairs of source and new target sequences to train the first Transformer of 2PT.}
\end{figure*}

{\fontfamily{pcr}\selectfont FastAlign} is an unsupervised word aligner that processes source and target sentences together and provides word-level alignments.  It is usable at training time but not for inference because it requires access to both sides and we only have the source side (at test time). As a solution, we used the alignments to create a training set and utilized it to train the first Transformer in 2PT. Figure \ref{fig:fastal} shows the input and output format in  {\fontfamily{pcr}\selectfont FastAlign} and how it helps generate training samples. 

As the figure demonstrates, given a pair of English--German sentences, word alignments are generated. In order to process the alignments, we designed rules to handle different cases: 
\begin{itemize}
  \item \textbf{One-to-Many Alignments}: We only consider the first target position in one-to-many alignments (see the figure).
  
  \item \textbf{Many-to-One Alignments}: Multiple source words are reordered together (as one unit while maintaining their relative positions with each other) using the position of the corresponding target word. 
  
  \item \textbf{No Alignment}: Words that do not have any alignments are skipped and we do not change their position 
  \item  We also ensure that no source word would be aligned with a position beyond the source sentence length. 
\end{itemize}

Considering these rules and what {\fontfamily{pcr}\selectfont FastAlign} generates, the example input sentence ``\textit{mr h\"{a}n@@ sch represented you on this occasion .}'' (in Figure \ref{fig:fastal}) is reordered to ``\textit{mr h\"{a}n sch you this represented .}'' @@ are auxiliary symbols added in between sub-word units during preprocessing. See Section \ref{exp} for more information. 

Using re-ordered sentences, the first Transformer in 2PT is trained to reorder the original source sentences, and the second Transformer, which is responsible for translation, receives the re-ordered source sentences and maps them to their target translations. Despite different data formats and different inputs/outputs, 2PT is still a pipeline that translates a source language to a target one through internal modifications that are hidden from the user. 

\subsection{Dynamic Position Encoding}
\begin{figure*}[t]
\centering
\includegraphics[scale=1]{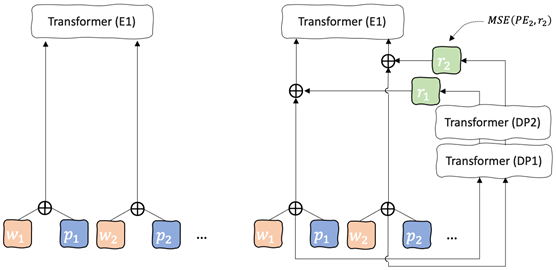}
\caption{\label{fig:dpe} The left-hand side is the original Transformer's architecture and the figure on the right is our proposed architecture. E1 is the first encoder layer of the ordinary Transformer and DP1 is the first layer of the DPE network.}
\end{figure*}

Unlike 2PT, the dynamic position encoding (DPE) method takes advantage of end-to-end training, while the source side still learns target reordering position information. It boosts the input of an ordinary Transformer's encoder with target position information, but leaves its architecture untouched, as illustrated in Figure \ref{fig:dpe}.

The input to DPE is a source word embedding ($w_i$) summed with sinusoidal position ($p_i$) encoding ($w_i \oplus p_i$). We refer to these embeddings as \textit{enriched embeddings}. Sinusoidal position encoding is part of the original design of Transformers and we assume the reader is familiar with this concept. For more details see \citet{vaswani2017attention}.

DPE is another neural network placed in-between the \textit{enriched embeddings} and the first encoder layer of the (translation) Transformer. In other words, the input to the DPE network is the embedding table of the Transformer, and its final layer outputs into the first encoder layer of the Transformer. Thus, the DPE network can be trained jointly with the Transformer using the original parallel sentence pairs. 

DPE processes enriched embeddings and generates a new form of them that is represented as $r_i$ in this paper, i.e. DPE($w_i \oplus p_i$) = $r_i$. DPE-generated embeddings are intended to preserve target-side order information about each word. In the original Transformer, the position of $w_i$ is dictated by adding $p_i$, but the original position of this word is not always the best one for translation; thus $r_i$ is defined to address this problem. If $w_i$ appears in the $i$\textit{-th} position but $j$ is its best position with respect to the target language, $r_i$ is supposed to learn information about the $j$\textit{-th} position and mimic $p_j$. Accordingly, the combination of $p_i$ and $r_i$ should provide $w_i$ with the pre-reordering information it requires to improve translation accuracy. 

In our design, DPE consists of two Transformer layers. We determined this number through an empirical study to find a reasonable balance between translation quality and resource consumption. These two layers are connected to an auxiliary loss function to ensure that the output of DPE is what we need for re-ordering.

This additional loss measures the mean squared error between the embeddings produced by DPE ($r_i$) and the supervising positions ($PE$) defined by {\fontfamily{pcr}\selectfont FastAlign} alignments. This learning process is simply formulated in Equation \ref{eq:sim}:
\begin{equation}\label{eq:sim}
\mathcal{L}_{order} = \frac{\sum_{i=1}^{|S|}MSE(PE_i, r_i)}{{|S|}}
\end{equation}
where $S$ is the source sequence length and $MSE()$ is the mean-square error function. The supervising position $PE_i$ is obtained by taking the target position associated with $w_i$ that was defined by {\fontfamily{pcr}\selectfont FastAlign} as described in Section \ref{twopass}. 

To clarify how $\mathcal{L}_{order}$ works, we use the aforementioned scenario as an example. We assume that the correct position for $w_i$ according to the {\fontfamily{pcr}\selectfont FastAlign} alignments is $j$, so $PE_i = p_j$ and we thus compute $MSE(p_j, r_i)$. Through this technique, we encourage the DPE network to learn pre-reordering in an end-to-end fashion and provide $w_i$ with position refinement information. 

The total loss function when training the entire model includes the auxiliary reordering loss function $\mathcal{L}_{order}$ summed with the standard Transformer loss $\mathcal{L}_{translation}$, as in Equation \ref{allloss}:
\begin{equation}\label{allloss}
    \mathcal{L}_{total} =  \lambda\times  \mathcal{L}_{translation} + (1-\lambda) \times \mathcal{L}_{order} 
\end{equation}
where $\lambda$ is a hyper-parameter that represents the weight of the reordering loss. $\lambda$ was determined by minimizing the total loss on the development set during training. 

\section{Experimental Study}\label{exp}
\subsection{Dataset}
To train and evaluate our models, we used the IWSLT-14 collection \citep{cettolo-etal-2012-wit3} and the WMT-14 dataset.\footnote{\url{http://statmt.org/wmt14/translation-task.html}} Our datasets are commonly used in the field, which makes our results easily reproducible. Our code is also publicly available to help other researchers further investigate this topic.\footnote{\url{https://github.com/jy6zheng/DynamicPositionEncodingModule}} The IWSLT-14 collection was used to study the impact of our model for the English--German (En--De), English--French (En--Fr), and English--Italian (En--It) pairs. We also reported results on the WMT dataset, which provides a larger training corpus. We know that the quality of NMT models vary in proportion to the corpus size, so these experiments provide more information to better understand our model. 

To prepare the data, sequences were lower-cased, normalized, and tokenized using the scripts provided by the Moses toolkit\footnote{\url{https://github.com/moses-smt/mosesdecoder}} \citep{koehn-etal-2007-moses} and decomposed into sub-words via Byte-Pair Encoding (BPE) \citep{sennrich2016neural}. The vocabulary sizes extracted for the IWSLT and WMT datasets were $32$K and $40$K, respectively. For the En--De pair of WMT-14, \textit{newstest2013} was used as a development set and \textit{newstest2014} was our test set. For the IWSLT experiments, our test and development sets were as suggested by \citet{Zhu2020Incorporating}. Table \ref{stat} provides the statistics of our datasets. 

\begin{table}[h]
\begin{tabular}{l l l l}\hline
Data &  Train & Dev  & Test  \\
 \hline
 WMT-14 (En$\rightarrow$De)  & 4.45M & 3k & 3k  \\
 IWSLT-14 (En$\rightarrow$De) & 160k & 7k  & 6k  \\
 IWSLT-14 (En$\rightarrow$Fr) & 168k & 7k & 4k  \\
 IWSLT-14 (En$\rightarrow$It) & 167k & 7k  & 6k \\
 \hline
\end{tabular}
\caption{\label{stat}The statistics of the datasets used in our experiments. Train, Dev, and Test stand for the training, development, and test sets, respectively.}
\end{table}

\subsection{Experimental Setup}
In the interest of fair comparisons, we used the same setup as \citet{chen2020explicit} to build our baseline for the WMT-14 En--De experiments. This baseline setting was also used for our DPE model and DPE-related experiments. Our models were trained on $8$ $\times$ V100 GPUs. Since our models rely on the Transformer's backbone, all hyper-parameters that were related to the main Transformer architecture, such as embedding dimensions, the number of attention heads, etc., were set to the default values proposed for \textit{Transformer Base} in \citet{vaswani2017attention}. Refer to the original work for detailed information.

For IWSLT experiments, we used a lighter architecture since the datasets were smaller than WMT. The hidden dimension was $256$ for all encoder and decoder layers, and a dimension of $1024$ was used for the inner feed-forward network layer. There were $2$ encoder and $2$ decoder layers, and $2$ attention heads. We found this setting through an empirical study to maximize the performance of our IWSLT models. 

For the WMT-14 En--De experiments, similar to \citet{chen2020explicit}, we trained the model for $300$K updates and used a single model obtained from averaging the last 5 checkpoints. The model was validated with an interval of $2$K on the development dataset. The decoding beam size was $5$. In the IWSLT-14 cases, we trained the models for 15,000 updates and used a single model obtained from averaging the last 5 checkpoints that were validated with an interval of 1000 updates. We evaluated all our models with detokenized BLEU \citep{papineni-etal-2002-bleu}.

\begin{table*}[h]
\centering
\begin{tabular}{l|l|l|l }
 \toprule
 & Model & Data type & BLEU Score\\
 \midrule
1 & Reordering Transformer & En $\rightarrow$En$_{reordered}$ & 35.21\\
 \hline
2 &Transformer Base & En$\rightarrow$De & 27.76\\
3 &+ fed with the output of reordering Transformer & En$_{reordered}\rightarrow$De & 21.96\\
4 & + fed with the output of {\fontfamily{pcr}\selectfont FastAlign} & En$_{reordered}\rightarrow$De & 31.82\\
 \bottomrule
\end{tabular}
 \caption{\label{tab:two-pass transformer}BLEU scores for the 2PT series of experiments.}
\end{table*}

\subsection{2PT Experiments}\label{twopass_experiment}
\begin{table*}[h]
\centering
\begin{tabular}{l | l |l }
 \toprule
 Model & \# Params &  En$\rightarrow$De (WMT)\\
 \midrule
 Transformer Base  \citep{vaswani2017attention} & 65.0 M & 27.30 \\
  + Relative PE \citep{shaw2018selfattention}  & N/A & 26.80  \\
  + Explicit Global Reordering Embeddings \citep{chen2020explicit}  &  66.5 M & 28.44 \\
  + Reorder fusion-based source representation
  \citep{chen2020explicit}  &  66.5 M & 28.55 \\
  + Reordering Embeddings (Encoder Only) \citep{chen-etal-2019-neural}  & 102.1 M & 28.03 \\
  + Reordering Embeddings (Encoder/Decoder) \citep{chen-etal-2019-neural}  &  106.8 M & 28.22 \\
 \midrule
 Transformer Base (our re-implementation) & 66.5 M & 27.78 \\
 Dynamic Position Encoding ($\lambda = 0.5$) & 72.8 M & 28.59\\ 
 \bottomrule
\end{tabular}
\caption{\label{tab:different layer number} A BLEU score comparison of DPE versus other peers.}
\end{table*}

Results related to the two-pass architecture are summarized in Table \ref{tab:two-pass transformer}. The reordering Transformer (Row 1) works with the source sentences and reorders them with respect to the order of the target language. This was a monolingual translation task with a BLEU score of $35.21$. This is a relatively low score for a monolingual setting which indicates how complicated the reordering problem is. Even dedicating a whole Transformer could not fully overcome the reordering problem. This finding also indicates that NMT engines can benefit from using an auxiliary module to handle order complexities. It is usually assumed that the translation engine should perform in an end-to-end fashion where it deals with all the reordering, translation, and other complexities via a single model at the same time. However, if we can separate these sub-tasks systematically and tackle them individually, there is a chance that we might be able to improve the overall quality. 

In Row 3, we used the information previously generated (in Row 1) and showed how a translation model performs when it is fed with reordered sentences. The BLEU score for this task was $21.96$, which was significantly lower than the baseline (Row 2). Order information was supposed to increase the overall performance, but we observe a degradation. This is because the first Transformer was unable to detect the correct order (due to the difficulty of this task). In Row 4, we fed the same translation engines with higher-quality order information (generated by {\fontfamily{pcr}\selectfont FastAlign}), and the BLEU score rose to 31.82. 

We cannot use {\fontfamily{pcr}\selectfont FastAlign} at test time but this experiment shows that our hypothesis on the usefulness of order information seems to be correct. Motivated by this, we invented DPE to better leverage order information, and these results are reported in the next section. 

\subsection{DPE Experiments}
Results related to DPE are reported in Table \ref{tab:different layer number}. According to the reported scores, DPE led to a +0.81 improvement in the BLEU score compared to \textit{Transformer Base}. To ensure that we evaluated DPE in a fair setup, we re-implemented the \textit{Transformer Base} in our own environment. This eliminated the impact of different factors and ensured that the gain was due to the design of the DPE module itself. We also compared our model to models discussed in the related literature such as that of \citet{shaw2018selfattention}, the reordering embeddings of \citet{chen-etal-2019-neural}, and the more recent explicit reordering embeddings of \citet{chen2020explicit}. Our model achieved the best score and we believe it was due to the direct use of order information. 

For the DPE architecture, we decided to have two layers (DP1 and DP2) as it produced the best BLEU scores on the development sets without imposing significant training overhead. One important hyper-parameter that directly affects DPE's performance is the position loss weight ($\lambda$). We ran an ablation study on the development set to adjust $\lambda$. Table \ref{tab:different lambda number} summarizes our findings. The best $\lambda$ value in our setting was $0.5$. This value provided an acceptable balance between translation accuracy and pre-reordering costs during training, and shows that the order information can be as important as other translation information. 

\begin{table}[h]
\centering
\begin{tabular}{l | l}
 \toprule
 Model & WMT'14 En$\rightarrow$ De \\
 \midrule
 Baseline   & 27.78 \\
 DPE ($\lambda = 0.1$) &  28.17  \\
 DPE ($\lambda = 0.3$) &  28.16 \\
 DPE ($\lambda = 0.5$) & \textbf{28.59} \\
 DPE ($\lambda = 0.7$)  & 27.98  \\
 \bottomrule
\end{tabular}
\caption{\label{tab:different lambda number} BLEU scores of DPE with different $\lambda$ values.}
\end{table}

\begin{table}[h!]
\centering
\begin{tabular}{l | l}
 \toprule
 Model & WMT'14 En$\rightarrow$De \\
 \midrule
 Baseline & 27.78 \\
 8E & 28.07\\
 10E & 28.54  \\
 \midrule
 DPE ($N=2$)  & 28.59 (\textbf{+ 0.81}) \\
 \bottomrule
\end{tabular}
\caption{\label{tab:different baseline} A BLEU score comparison of DPE with the baseline Transformer models plus additional encoder layers (8E for 8 encoder layers and 10E for 10 encoder layers)}
\end{table}

The design of our Transformer (Transformer Base + DPE) might raise the concern that incorporating pre-reordering information or defining an auxiliary loss might not be necessary. One might suggest that if we use the same amount of resources to increase the Transformer Base's encoder parameters, we should obtain competitive or even better results than the DPE-enhanced Transformer. To address this concern, we designed another experiment that increased the number of parameters/layers in the \textit{Transformer Base} encoder to match the number in our model's parameters. Results related to this experiment are shown in Table \ref{tab:different baseline}. 

The comparison of DPE with the different extensions of the Transformer Base, namely 8E (8 encoder layers) and 10E (10 encoder layers), demonstrated that the increase in BLEU was due to the position information provided by DPE rather than the additional parameters of the DPE layers. In 8E, we provided the same number of additional parameters as the DPE module adds, but experienced less gain in translation quality. In 10E, we even doubled the number of additional parameters to surpass the number of parameters that DPE uses, and yet the DPE extension with 8 encoding layers (two for pre-reordering and six from the original translation encoder) was still superior. This reinforces the idea that our DPE module improves translation accuracy by injecting position information alongside the encoder input. 

\begin{table*}[ht]
\centering
\begin{tabular}{l | l |l |l}
 \toprule
Model & En$\rightarrow$De & En$\rightarrow$Fr & En$\rightarrow$It \\
 \midrule
 Transformer & 26.42 & 38.86 & 27.94 \\
 DPE-based Extension & 27.47 ($\uparrow$ 1.05)& 39.42 ($\uparrow$ 0.56) & 28.35 ($\uparrow$ 0.41)\\
 \bottomrule
\end{tabular}
\caption{\label{tab:different language} BLEU results for different IWSLT-14 Language pairs.}
\end{table*} 

\subsection{Experimental Results on Other Languages}
In addition to the previously reported experiments, we evaluated the DPE model on different IWSLT-14 datasets of English--German (En--De), English--French (En--Fr), and English--Italian (En--It). After tuning with different position loss weights on the development set, we determined $\lambda = 0.3$ to be ideal for this setting. The results in Table \ref{tab:different language} show that with DPE, the translation accuracy improved for different settings and the improvement was not unique to the En--De WMT language pair.

Our DPE architecture works with a variety of language pairs of different sizes and this increases our confidence in the beneficial effect of order information. It is usually hard to show the impact of auxiliary signals in NMT models and this could be more difficult with smaller datasets, but our IWSLT results are promising. Accordingly, it would not be unfair to claim that DPE is useful regardless of the language and dataset size.  

\section{Conclusion and Future Work}
In this paper, we first explored whether Transformers would benefit from order signals. Then, we proposed a new architecture, DPE, that generates embeddings containing target word position information to boost translation quality. 

The results obtained in our experiments demonstrate that DPE improves the translation process by helping the source side learn target position information. The DPE model consistently outperformed the baselines of related literature. It also showed improvements with different language pairs and dataset sizes. 
\subsection{Future Work}
Our experiments can provide the groundwork for further exploration of dynamic position encoding in Transformers. First, we acknowledge that there are some extensions to our current work. Additional rules can be designed to handle different cases of word alignments generated by {\fontfamily{pcr}\selectfont FastAlign}. For example, cases such as Many-to-Many alignments and multi-word expressions are also frequently found in written text. Another possible extension would be to investigate more precise alignment tools in addition to {\fontfamily{pcr}\selectfont FastAlign}. However, it is important to note that we did not heavily invest in linguistic preprocessing because it requires too many resources. Extremely precise preprocessing might not be necessary as neural models are expected to still solve problems with limited access to domain information. When considering other alignment tools, we must also consider the efficiency and scalability of our solution. 

Finally, we plan to explore injecting order information into other language processing models through DPE or a similar mechanism. Such information seems to be useful for tasks such as dependency parsing or sequence tagging.  
\normalem
\bibliography{anthology,custom}
\bibliographystyle{acl_natbib}

\end{document}